\documentclass[conference]{IEEEtran}
  \IEEEoverridecommandlockouts
  \usepackage{cite}
  \usepackage{amsmath,amssymb,amsfonts}
  \usepackage{algorithmic}
  \usepackage{ulem}
  \usepackage{amsmath}
  \usepackage{amsthm}
  \usepackage{graphicx}
  \usepackage{subfigure}
  \usepackage{color}
  \usepackage{graphicx}
  \usepackage{textcomp}
  \usepackage{multirow}
  \usepackage{booktabs}
  \usepackage{threeparttable}
  \usepackage{xcolor}
  
  \usepackage[marginal]{footmisc}

  \def\BibTeX{{\rm B\kern-.05em{\sc i\kern-.025em b}\kern-.08em
  		T\kern-.1667em\lower.7ex\hbox{E}\kern-.125emX}}
  \begin{document}
  	
  	\title{Single Image Deraining via Feature-based Deep Convolutional Neural Network\\
  		{\footnotesize \textsuperscript{}}
  	}
  	
  	\author{\IEEEauthorblockN{Chaobing Zheng}
  		\IEEEauthorblockA{School of Information \\ Science and Engineering, \\
  			Wuhan University of \\ Science and Technology,\\
  			Wuhan, China \\
  			zhengchaobing@wust.edu.cn}
  		\and
  		\IEEEauthorblockN{Jun Jiang {*}}
  		\IEEEauthorblockA{School of Information \\ Science and Engineering, \\
  			Wuhan University of \\ Science and Technology,\\
  			Wuhan, China \\
  			jiangjun85@wust.edu.cn}
  		\and
  		\IEEEauthorblockN{Wenjian Ying }
  		\IEEEauthorblockA{College of Weapon, \\ Naval University of Engineering,\\
  			Wuhan, China \\
  			842812952@qq.com}
  		\and
  		\IEEEauthorblockN{Shiqian Wu \dag}
  		\IEEEauthorblockA{School of Information  \\ Science and Engineering, \\
  			Wuhan University of \\ Science and Technology,\\
  			Wuhan, China \\
  			shiqian.wu@wust.edu.cn}
  	}
  	
  	\maketitle
\maketitle

\begin{abstract}
It is challenging to remove rain-steaks from a single rainy image because the rain steaks are spatially varying in the rainy image. Although the CNN based methods have reported promising performance recently, there are still some defects, such as data dependency and insufficient interpretation. A single image deraining algorithm based on the combination of data-driven and model-based approaches is proposed. Firstly, an improved weighted guided image filter (iWGIF) is used to extract high-frequency information and learn the rain steaks to avoid interference from other information through the input image. Then, transfering the input image and rain steaks from the image domain to the feature domain adaptively to learn useful features for high-quality image deraining. Finally, networks with attention mechanisms is used to restore high-quality images from the latent features. Experiments show that the proposed algorithm significantly outperforms state-of-the-art methods in terms of both qualitative and quantitative measures. 
\end{abstract}
\begin{IEEEkeywords}
	single image rain removal, weighted guided image filtering,  data-driven, model-based
\end{IEEEkeywords}

\footnotetext{ *  contributed equally, $\dag$  corresponding author. This work is supported by Nature Science Foundation of Hubei Province (Grant No.2022CFB676)} 

\section{Introduction}
As the most common severe weather condition, the impact of rain will reduce the visual quality of images and seriously affect the performance of outdoor visual system. Under rainy conditions, the rain steaks will not only produce a fuzzy effect in the image, but also cause haze due to light scattering, low visibility of the scene and occlusion of the background scene. The contrast and color of the target in the image will be attenuated to varying degrees, resulting in unclear expression of the background information (i.e. the target image), which makes some video or image systems unable to work normally \cite{zheng22}. Therefore, it is necessary to eliminate the impact of rain on the image scene. In fact, image rain removal has always been an important part of image restoration and computer vision research, which is mainly used in video surveillance, automatic driving and other fields.

Single image de-raining is more challenging than video-based de-raining because there is less available information. It has attractted many researchers's attention to solve the problem of restoring rainy images. The existing rain steaks removal methods for a single image can generally be divided into two categories: model-based methods \cite{Liu, Li2016, 1kang2012, zhu} and data-driven methods \cite{ Zhang1, Zhang2, 1fu2017, 1yang2017, 1li2018}. As we known, the model-based rain removal algorithm is not robust enough, with excessive processing or retention of rain steaks, and is not universal enough. Data-driven rain removal algorithm is very popular. However, there are defects in data dependence and insufficient interpretability.

A novel removal algorithm by combining data-driven and model-based approaches is proposed in this paper. The proposed algorithm embeds deep learning into a prior model, making full use of deep learning to enhance the robustness of the algorithm. The priori model is used to enhance the interpretability of the algorithm. The two algorithms are combined to complement each other \cite{Zheng2022, Li2022, Zheng2023}. Experimental results show that the proposed deraining algorithm outperforms a few existing deraining algorithms. It also proves that the algorithm combining deep learning and model-based can complement each other.

The rest of this paper is organized as below. The relevant works is proposed in Section \ref{Relevant}. The proposed deraining algorithm is presented in Section \ref{algorithm}. Experimental results are provided in Section \ref{Experiments} to verify the proposed algorithm. Finally, conclusion remarks are drawn in Section \ref{conclusion}

\section{Relevant Works}
\label{Relevant}
Rain removal has attracted many researchers's attention to solve the problem of restoring rain images. These methods fall into two categories: model-based methods and data-driven methods.

{\it Model-driven methods}: Some priors have been employed to remove rain from single images. They assume that rain-streaks $R$ are sparse and in similar directions. They decompose the raining image $O$ into two layers, the rain-free background scene $B$ and the rain-streaks layer $R$. Zhu et al. \cite{zhu} first detected rain-dominant regions and then the detected regions were utilized as a guidance image to help separate rain-streaks $R$ from the background layer $B$. Liu et al. \cite{Liu} decomposed the rainy image into low-frequency and high-frequency through low-pass filtering, captures the rain-free image details from the high-frequency part, superimposes the rain-free image details on the low-frequency part, and further alleviates the blurring caused by rain steaks by using the dark channel prior algorithm, but the rain marks of the derainy image is too obvious. Li et al. \cite{Li2016} proposed a rain removal method based on the Gaussian mixture model, which achieved good results in many cases. However, there are still rain line residues in the rain removal images and it is difficult to accurately estimate the Gaussian mixture model. Due to the effect of air resistance and wind, the shape and direction of movement of rain are random during the falling. It is difficult for model-driven algorithms to capture rules, so the effect of rain removal may be not ideal.

{\it Data-driven methods}: Data-driven based single image deraining algorithms were popular. Fu et al. \cite{1fu2017} learned the mapping relationship between rainy and clean detail layers, and the predicted detail layer was added into a low-pass filtered base layer for removing rain-streaks. Yang et al. \cite{1yang2017} proposed a deep recurrent dilated network to jointly detect and remove rain-streaks. Zhang et al. \cite{Zhang1} take the rain density into account and present a multi-task CNN for joint rain density estimation and deraining. Later,  they further improve their work by propose a conditional generative adversarial network for rain-streaks removal in \cite{Zhang2}. Li et al. \cite{1li2018} introduced a novel multi-stage deep learning based single image deraining algorithm. Different $\alpha$ values are assigned to various rain-streak layers according to their properties at each stage, and then a recurrent neural network was incorporated to preserve the useful information in previous stages and benefit the residual prediction in later stage. Data-driven relies on massive amounts of data and is superior to model-driven algorithms in terms of performance. However, even data-driven methods struggle to capture regularity due to the random nature of rainfall.

In this paper, a novel single-image rain removal method by combining data-driven and model-based is designed. The proposed algorithm is based on an observation that prior knowledge can assist data-driven better targeted learning. Since both rain and noise are random, they exist in high frequency components. The high-frequency part can be obtained by an edge-preserving filter such as \cite{1LiIEEETIP2015, 1chen2019}, and the rain steak can be learned directly from the high frequency information rather than the input image to avoid interference from other information. Then, transforming the input image and rain steaks image from the image domain to the feature domain, and adaptively inging the useful features to remove rain \cite{DWDN}. Finally, networks with attention mechanisms is used restore high-quality images from the latent features, which will inevitably cause errors due to inaccurate model. Experimental results validate the prior is indeed very helpful for the single image deraining. The proposed algorithm is useful for autonomous navigation in raining conditions.inging

\begin{figure*}[htb]
	\centering
	\includegraphics[width=0.98\textwidth]{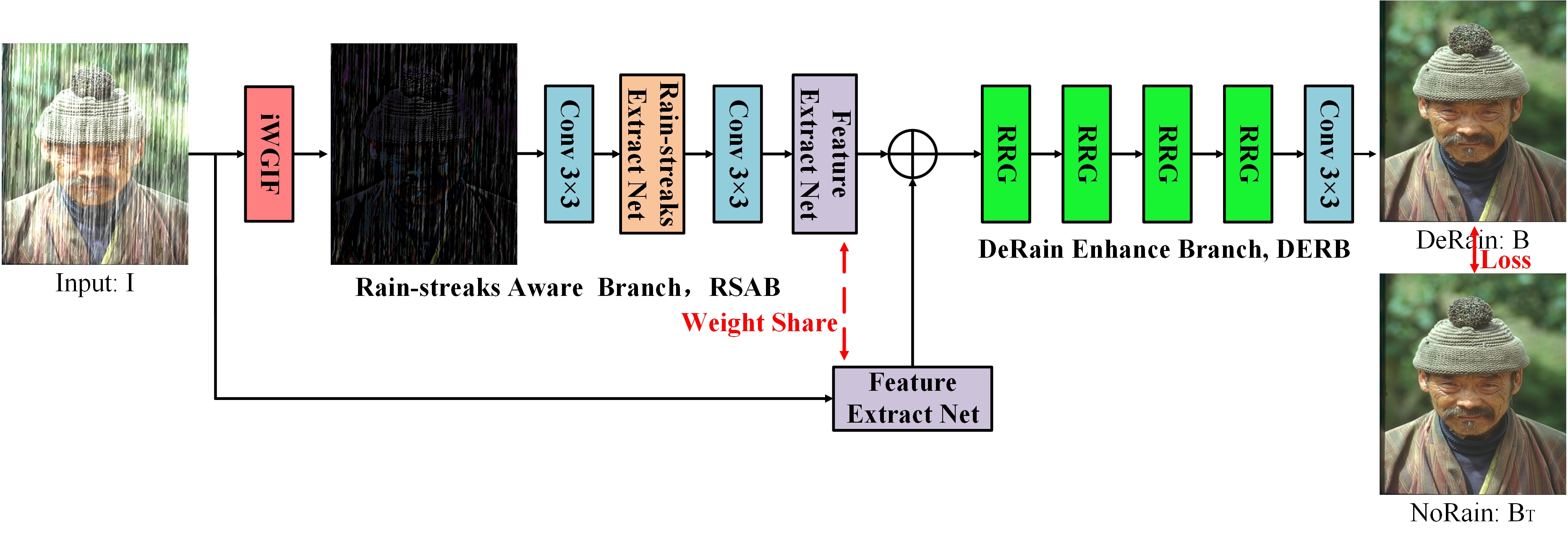}
	\caption{The proposed framework for single-image rain removal. $I$ is a rainy image, $B$ is a restored image, and $B_T$ is the ground truth image of $B$.}
	\label{Fig1}
\end{figure*}

\section{The Proposed Deraining Algorithm}
\label{algorithm}

\subsection{Framework of The Proposed Algorithm}
\label{fundation}

The widely used rainy image model is expressed as \cite{1kang2012}:
{\small \begin{equation}
	\label{eq1}
	I(p)=B(p)+S(p)\\
	\end{equation}}
where $I$ is a rainy image with rain-streaks; $B$ is the background layer; $S$ is the rain-streak layer; and $p$ is a pixel.  
Single image deraining is to restore the image $B$ from the rainy image $I$. However, removing rain from a single image is an ill posed problem. Especially, when the structure and direction of the object are similar to the rain steaks, even though the DCNN has a strong learning ability, it is also difficult to remove the rain steaks and maintain the details at the same time.

A new single image rain-streaks removal algorithm is proposed in this paper is by combining data-driven and model-based method. As shown in Fig. \ref{Fig1}, based on the model of $I(p)=B(p)+S(p)$, and the prior knowledge that rain steaks exist in high-frequency information, due to the disadvantage of DCNN for learning high-frequency information, directly using the rainy image to learn rain steaks is not ideal. Therefore, an improved weighted guided image filter (iWGIF) is adopted to obtain high frequency components which mainly contain noise, rain-streaks, etc. A Rain-streaks Extract Net is proposed to obtain rain steaks. Due to the limitation of prior knowledge, the high-frequency information obtained cannot include all rain steaks, the learned rain steaks may have inaccurate information, which needs further enhancement. Therefore, converting the rainy image and rain steaks from the image domain to the feature domain can reduce the interference of noise and other useless information, make full use of effective information, and improve the network training efficiency. Through the feature extraction network, the rainy image and rain steaks are converted from the image domain to the feature domain, and obtain the derain feature, as shown in Eq. \ref{eq2}.
{\small \begin{equation}
	\label{eq2}
	F(I(p)) = F(B(p)) + F(S(p))\\
	\end{equation}}
where $F( \cdot )$ represents the feature extraction function btained through feature extraction network. Finally, DeRain Enhance Branch (DERB) is used for image enhancement and restoration. The combination of prior knowledge and deep learning enhances the robustness and physical interpretability of the algorithm.

\begin{figure*}[htb]
	\centering
	\includegraphics[width=0.95\textwidth]{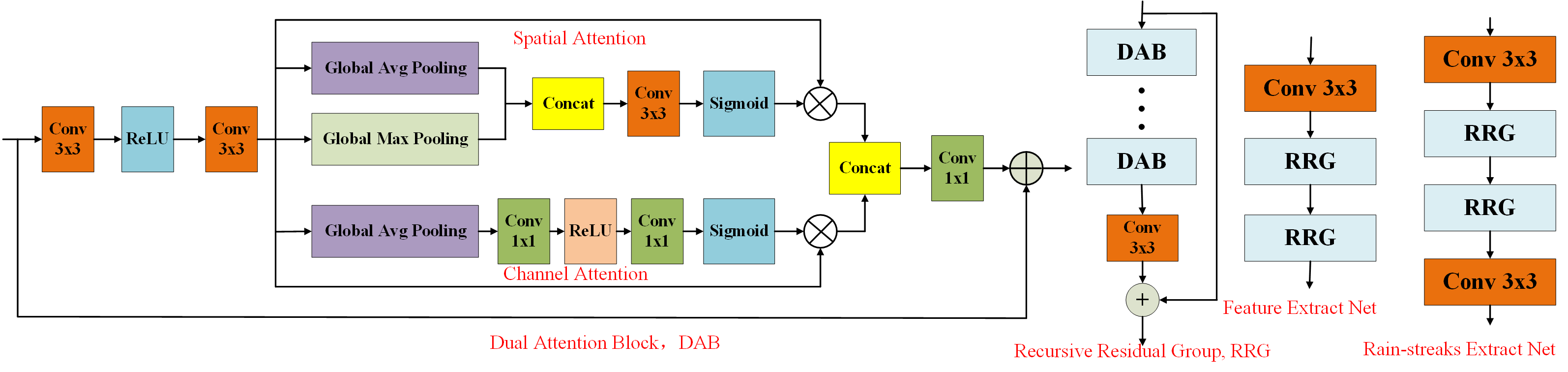}
	\caption{ Recursive residual group (RRG), Feature Extract Net and Rain-sreaks Extract Net, each RRG contains multiple dual attention blocks (DAB). Each DAB contains spatial and channel attention modules.}
	\label{Fig2}
\end{figure*}

\subsection{Improved Weighted Guided Image Filtering \cite{Li2022}}

In this subsection, an iWGIF is provided by using the edge-aware weighting in \cite{1LiIEEETIP2015} to improve the WAGIF in \cite{1chen2019}. The objective is to reduce the sensitivity of the WAGIF with respect to the regularization parameter \cite{Li2022}.

Let $I$ be an image to be processing and $G$ be a guidance image.  Let $\Omega_{\zeta}(p)$ be a square window centered at the pixel
$p$ of a radius $\zeta$. Same as  the GIF, WGIF, GGIF and WAGIF,  $I(p)$ is assumed to be a linear transform of the guidance image $G(p)$ in the window $\Omega_{\zeta}(p')$:
\begin{equation}
\label{linearmodel} I(p)=a_{p'}G(p)+b_{p'}, \forall p\in
\Omega_{\zeta}(p'),
\end{equation}
where $a_{p'}$ and $b_{p'}$ are two constants in the
window $\Omega_{\zeta}(p')$.

It should be pointed out that the guidance image $G$ and the image to be processed $I$ can be the same. The optimal values of $a_{p'}$ and $b_{p'}$ are derived by minimizing
a cost function $E(a_{p'},b_{p'})$ which is defined as
\begin{eqnarray}
\label{Eap''bp''old}
\sum_{p\in
	\Omega_{\zeta}(p')}[\Gamma^{G}_{p'}(a_{p'}G(p)+b_{p'}-I(p))^2+\lambda a_{p'}^2],
\end{eqnarray}
where $\lambda$ is a regularization parameter which can be used to penalize a large $a_{p'}$.

Similarly to the WGIF, the edge-aware weighting $\Gamma^{G}_{p'}$ is defined  by using local variances of $3\times 3$ windows of all pixels as follows:
\begin{eqnarray}
\label{w(p)}
\Gamma^{G}_{p'}=\frac{1}{M}\sum_{p=1}^{N}\frac{\sigma^2_{G,1}(p')+\varepsilon}{\sigma^2_{G,1}(p)+\varepsilon},
\end{eqnarray}
where $M$ is the total number of pixels in the image $I$, and  $\varepsilon$ is a small constant. Certainly, there are many different methods to compute the edge-ware weighting \cite{1LiIEEETIP2015}. Different edge-aware weighting could be selected for different applications.

By solving the optimization problem (\ref{Eap''bp''old}), the optimal values of $a_{p'}$ and $b_{p'}$ are computed as
\begin{align}
a_{p'}&=\frac{\Gamma^{G}_{p'}\mbox{cov}_{I,G,\zeta}(p')}{\Gamma^{G}_{p'}\sigma^2_{G, \zeta}(p')+\lambda},\\
b_{p'}&=\mu_{I,\zeta}(p')-a_{p'}\mu_{G,\zeta}(p'),
\end{align}
where the operation $\odot$ is the element-by-element product of two matrices. $\mbox{cov}_{I,G,\zeta}(p')$ is
\begin{align}
\mbox{cov}_{I,G,\zeta}(p')=\mu_{G\odot I, \zeta}(p')-\mu_{G,\zeta}(p')\mu_{I,\zeta}(p'),
\end{align}
and $\mu_{G, \zeta}(p')$, $\mu_{I, \zeta}(p')$, and  $\mu_{G\odot I, \zeta}(p')$ are the mean values of $G$, $I$ and $G\odot I$ in the window $\Omega_{\zeta}(p')$,  respectively.

Instead of using the averaging method in \cite{1he2013,1LiIEEETIP2015}, an optimal based method is adopted to compute the final value of $I(p)$ as
\begin{equation}
\min_{I(p)}\{\sum_{p'\in \Omega_{\zeta}(p)}W_{p'}(I(p)-a_{p'}G(p)-b_{p'})^2\},
\end{equation}
where $W_{p'}$  is a weighting factor and it is given as \cite{1chen2019}
\begin{align}
W_{p'}&=\exp^{-\frac{1}{|\Omega_{\zeta}(p')|}{\displaystyle \sum_{p\in \Omega_{\zeta}(p')}}\frac{(a_{p'}G(p)+b_{p'}-I(p))^2}{\eta}}+0.001,
\end{align}
$\eta$ is a small positive constant, and $|\Omega_{\zeta}(p')|$ is the cardinality of the set $\Omega_{\zeta}(p')$.

The optimal solution is computed as
\begin{equation}
I^*(p)=\bar{a}_p^WG(p)+\bar{b}_p^W,
\end{equation}
where $\bar{a}_p$ and $\bar{b}_p$ are computed as
\begin{align}
\bar{a}_p^W& =\frac{1}{W^{sum}_p}\sum_{p'\in \Omega_{\zeta}(p)}W_{p'}a_{p'},\\
\bar{b}_p^W& =\frac{1}{W^{sum}_p}\sum_{p'\in \Omega_{\zeta}(p)}W_{p'}b_{p'},
\end{align}
and $W_p^{sum}$ is
\begin{align}
W_p^{sum}&=\sum_{p'\in \Omega_{\zeta}(p)}W_{p'}.
\end{align}

It can be easily derived that
\begin{align}
\nonumber
&\frac{1}{|\Omega_{\zeta}(p')|}\sum_{p\in \Omega_{\zeta}(p')}(a_{p'}G(p)+b_{p'}-I(p))^2\\\nonumber
=&a_{p'}^2\sigma_{G,\zeta}^2(p')-2a_{p'}\mbox{cov}_{I,G,\zeta}(p')+\sigma_{I,\zeta}^2(p')\\
=&\sigma_{I,\zeta}^2(p')-a_{p'}^2(\sigma_{G,\zeta}^2(p')+\frac{2\lambda}{\Gamma_{p'}^G}).
\end{align}

Therefore, same as the GIF in \cite{1he2013}, WGIF in \cite{1LiIEEETIP2015} and WAGIF in \cite{1chen2019}, the complexity of the iWGIF is $O(M)$ for an image with $M$ pixels.

The iWGIF is adopted to decompose the rainy image $I$ into base layer and detail laye, the guidance image $G$ is the same as the rainy image $I$. The detail layer mainly contains high-frequency information, and most rain steaks are in the detail layer. Using the detail layer to learn the rain is more direct, avoiding other interference, and improving the learning efficiency of the network.

\subsection{Feature-based deep convolutional neural network}
Due to the limited modeling ability of prior knowledge, high-frequency information cannot cover all raindrops. In addition, DCNN is not good at obtaining high-frequency information, the learned rain steaks have incorrect information, it is necessary to enhance de-rainy image. A feature extract network is proposed, which transforms the rainy image and rain steaks from the image domain to the feature domain to adaptively learn useful features for high-quality image deraining, and enhances the efficiency of the network. Fig. \ref{Fig1} and \ref{Fig2} summarizes the pipeline of our algorithm. To make efficient use of useful features, the CNN used in the proposed framework is on top of the DeNoiseNet in \cite{CycleISP}, which achieves outstanding performance in denoising. The recursive residual group (RRG) in DeNoiseNet is shown in Fig. \ref{Fig2}. The RRG contains Dual Attention Blocks (DAB) and each DAB performs both spatial and channel attention operations. The feature extract network contains two RRGs and one $3*3$ convolution, which can suppress the less useful features and only allow the propagation of more informative ones. The rain-streaks extract network contains two $3*3$ convolution and two RRGs, which can pay more attention to the learning of rain steaks under the effect of attention mechanism. 

To restore high-quality images from the latent features $F(I(p)) - F(S(p))$, deRain enhance branch (DERB) is adopted. The DERB contains four RRGs and one $3*3$ convolution. By using the DERB, further feature extraction and recovery, the experiment also proves the importance of DERB.

Loss functions play an important role in training the CNN from $N$ pairs of images $\{(I,B_T)\}$. Here, $I$ is a rainy image and $B_T$ is a  rain-free image. To better regularize the network, we apply an $L1$ loss functions. The loss function is given as:
\begin{equation}
\label{lrlr}
L_r =\frac{1}{N} \sum_p  |B_T - B|,
\end{equation}

\section{Experiments}
\label{Experiments}
In this section, we first introduce the details of the implementation, then evaluate the effectiveness of the proposed method on the existing deraining datasets, compare the proposed deraining method  with several state-of-the-art single-image deraining methods. Finally, ablation experiments are designed to prove the contribution of each module. The details are as follows:

\subsection{Implementation Details}
To verify the proposed derain method, we conduct the comparison and analysis experiments on synthetic rain image Rain100L \cite{DJRain}, Rain100H \cite{DJRain} and Rain1400 \cite{DDRain}, and use SSIM and PSNR to evaluate the performance.

Each RRG contains two DAB. The number of RRG of SERB is set as four. We randomly crop a $128*128$ patch from each input during training. The proposed network is trained using the proposed loss functions and Adam optimizer with ${{\beta}_1}=0.9$ and ${{\beta}_2}=0.99$. We set the batch size to $8$. The learning rate is initially set to $2*10^{-4}$ and then decreased using a cosine annealing schedule. All the experiments are implemented using PyTorch on NVIDIA GP100 GPUs.

\subsection{Comparison with Existing Algorithms}

\begin{figure*}[htb!]
	\centering
	\includegraphics[width=0.95\textwidth]{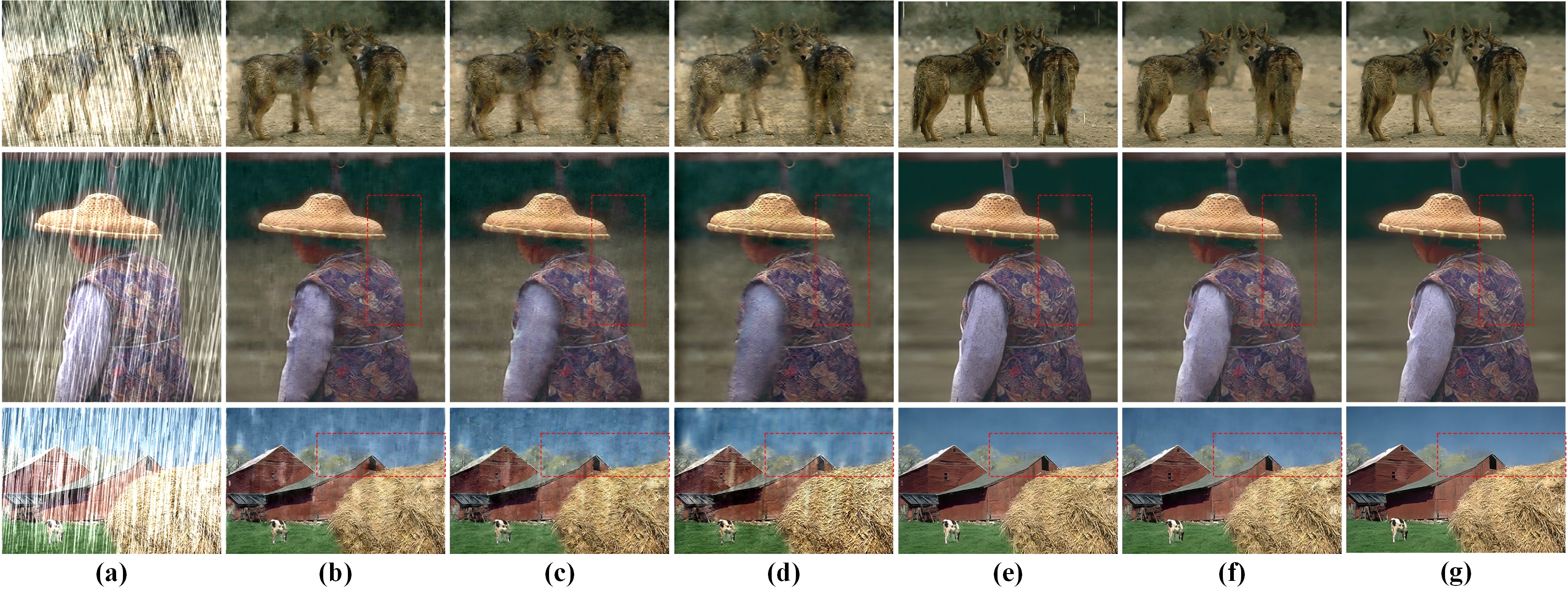}
	\caption{Comparison of different deraining algorithms on synthetic rainy images. From left to right, the rainy images and the restored images by  \cite{DJRain},  \cite{DDRain},  \cite{LPNet},  \cite{RCDNet}, \cite{VRGNet} and the proposed method, respectively. }
	\label{Fig4}
\end{figure*}

The proposed algorithm is compared with five state-of-the-art deraining algorithms in \cite{DJRain},  \cite{DDRain},  \cite{LPNet},  \cite{RCDNet}, \cite{VRGNet} on synthetic rainy dataset. These algorithms were published at top conferences in computer vision. As shown in Fig. \ref{Fig4},  the algorithms in \cite{VRGNet, RCDNet} and the proposed algorithm can be adopted to reduce rain-streaks better than the algorithms in \cite{DJRain, DDRain, LPNet}. There are still visible artifact in the restored images by the algorithm in \cite{VRGNet, RCDNet}, as shown in the red box. The \cite{DDRain} also uses high-frequency information to learn rain steaks. However, DCNN is not good at learning high-frequency information, and there are obvious traces. 

Besides the subjective evaluation, the objective quality metrics including the SSIM and PSNR are adopted to further compare the proposed algorithm with those in \cite{DJRain},  \cite{DDRain},  \cite{LPNet},  \cite{RCDNet}, \cite{VRGNet}.  The average SSIM and PSNR are shown in Tables \ref{tabfusion1}-\ref{tabfusion3}. The proposed algorithm can achieve higher PSNR and SSIM than others.

\begin{table}[htb!]
	\vspace{-4mm}
	\caption{SSIM and PSNR Of Different Algorithms for Rain100L}
	\centering
	{\scriptsize\begin{tabular}{ c|ccccccccc }
			\hline
			& \cite{DJRain}&\cite{DDRain}&\cite{LPNet}& \cite{RCDNet}& \cite{VRGNet} & Ours \\
			\hline
			SSIM &0.965 &0.948     &0.952   &\textbf{0.984}   &0.98     &\textbf{0.984}   \\
			PSNR &34.32 &32.94     &32.15   &\textbf{38.58}   &38.21    &38.43  \\
			\hline	
	\end{tabular}}
	\label{tabfusion1}
\end{table}

\begin{table}[htb!]
	\vspace{-4mm}
	\caption{SSIM and PSNR Of Different Algorithms for Rain100H}
	\centering
	{\scriptsize\begin{tabular}{ c|ccccccccc }
			\hline
			& \cite{DJRain}&\cite{DDRain} &\cite{LPNet}& \cite{RCDNet}& \cite{VRGNet} & Ours \\
			\hline
			SSIM &0.786 &0.751   &0.760  &0.886  &0.855    &\textbf{0.903}   \\
			PSNR &24.37 &24.56   &21.79  &28.81  &27.62    &\textbf{29.53}  \\
			\hline	
	\end{tabular}}
	\label{tabfusion2}
\end{table}

\begin{table}[htb!]
	\vspace{-4mm}
	\caption{SSIM and PSNR Of Different Algorithms for Rain1400}
	\centering
	{\scriptsize\begin{tabular}{ c|ccccccccc }
			\hline
			& \cite{DJRain}&\cite{DDRain} &\cite{LPNet}& \cite{RCDNet}& \cite{VRGNet} & Ours \\
			\hline
			SSIM &0.876   &0.864   &0.877  &0.909  &0.908   &\textbf{0.912}   \\
			PSNR &28.37   &28.19   &28.21  &30.59  &30.6    &\textbf{31.04}  \\
			\hline	
	\end{tabular}}
	\label{tabfusion3}
\end{table}

\begin{table*}[htb]
	\begin{center}
		\centering
		\caption{Ablation study on key components of the proposed exposure interpolation framework on Rain100H ($\uparrow$: larger is better)}
		{\small \tabcolsep12pt\begin{tabular}{c|ccc|cc}
				\hline 		
				Case  & iWGIF  & Feature Extract Net  & DERB & SSIM ($\uparrow$)    & PSNR ($\uparrow$) \\
				\hline
				1 & N   & Y  & Y  &  0.885   & 28.64  \\
				2 & Y   & N  & Y  &  0.885   & 28.58 \\
				3 & Y   & Y  & N  &  0.802   & 25.22 \\
				4 & Y   & Y  & Y  &  0.903   & 29.53\\
				\hline
		\end{tabular}}
		\label{ta2b}
	\end{center}
\end{table*}

\begin{figure*}[htb!]
	\centering
	\includegraphics[width=0.98\textwidth]{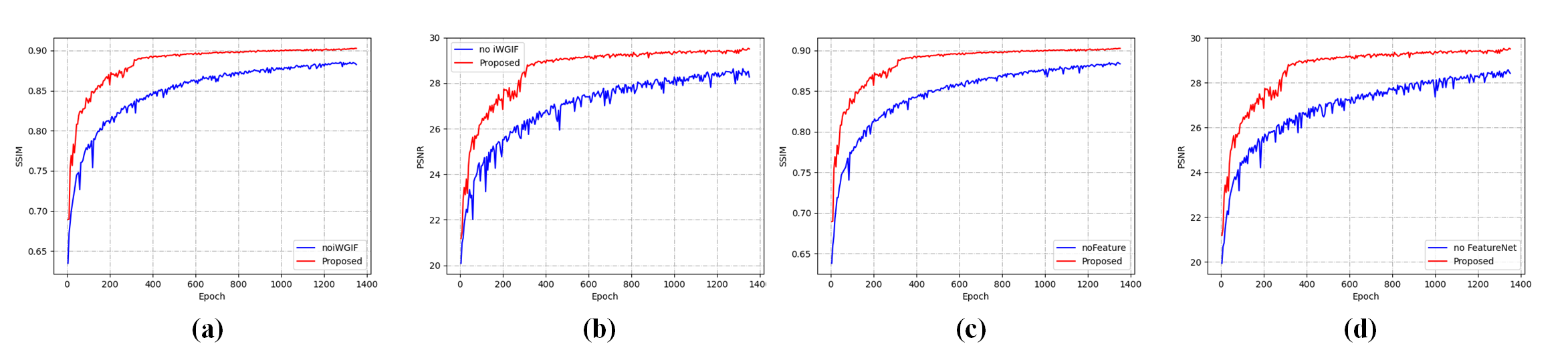}
	\caption{(a) and (c) Comparisons of PSNR between the modes with and without the iWGIF and Feature Net for Rain100H. (b) and (d) Comparisons of SSIM between the modes with and without the iWGIF and Feature Net for Rain100H. }
	\label{Fig5}
\end{figure*}

\subsection{Ablation Study}
\begin{figure}[htb!]
	\centering
	\includegraphics[width=0.48\textwidth]{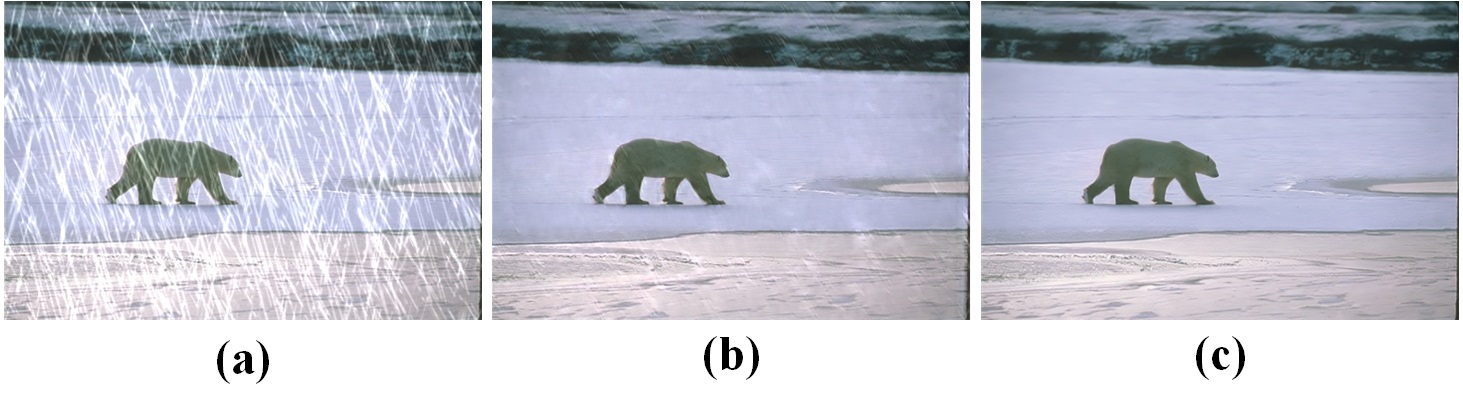}
	\caption{Comparison of results with and without DERB.}
	\label{Fig6}
\end{figure}

Key components of the proposed framework are: 1) iWGIF, 2) DERB,  3) Feature Extract Net.  Their performances are evaluated in this subsection. As illustrated in Table \ref{ta2b},  the mode with the iWGIF is more stable than the mode without iWGIF, and can achieve higher PSNR and SSIM values in different epoches. It also proves the performance of feature extraction network as shown in Fig. \ref{Fig5}. The feature extraction network can adaptively use useful features for image enhancement and improve network training efficiency. The results with and without EDRB are shown in Fig. \ref{Fig6}, the DERB can avoid the inaccuracy of modeling and further enhance the latent features.

\section{Conclusion Remarks and Discussions}
\label{conclusion}
A novel single image deraining algorithm by combinating data-driven and model-based methods is proposed in this paper. By using high-frequency information extracted by iWGIF to learn rain steaks, the training efficiency of CNN is improved. Combined with the rain removal model, the input image and rain steaks are converted from the image domain to the feature domain, which can use the useful information adaptively. Finally, the latent features are enhanced and restored through the network with dual attention mechanism. Experimental results show that the proposed algorithm could outperform several existing single image de-raining algorithms.

For other low-level visual processing tasks such as dehazing and illumination enhancement, the prior knowledge can also be incorporated into data-driven approaches. Besides the prior knowledge, it is also important to combine model-based and data-driven methods for the low-level visual processing. The proposed algorithm can also be applied to study autonomous navigation in raining conditions. All these problems will be studied in our future research.

\end{document}